\documentclass[10pt, a4paper]{article}

\usepackage{lrec-coling2024} % this is the new style
\usepackage{tcolorbox}
\usepackage[noend]{algpseudocode}
\usepackage{algorithm,bm}
\usepackage{amsmath}
\usepackage{booktabs}
\usepackage{multirow}
\usepackage{fancyhdr}

\usepackage{array,etoolbox}
\preto\tabular{\setcounter{magicrownumbers}{0}}
\newcounter{magicrownumbers}

\newenvironment{myfont}{\fontfamily{ptm}\selectfont}{\par}

\title{A Dual-View Approach\\ to Classifying Radiology Reports by Co-Training}

\name{Yutong Han$^1$, Yan Yuan$^2$, Lili Mou$^{1,3}$} 

\address{$^1$Dept. Computing Science, Alberta Machine Intelligence Institute (Amii), University of Alberta \\ $^2$School of Public Health, University of Alberta \\ 
$^3$Canada CIFAR AI Chair, Amii\\
yhan22@ualberta.ca,\quad yyuan@ualberta.ca,\quad doublepower.mou@gmail.com\\}

\abstract{
Radiology report analysis provides valuable information that can aid with public health initiatives, and has been attracting increasing attention from the research community. In this work, we present a novel insight that the structure of a radiology report (namely,  the \textit{Findings} and \textit{Impression} sections) offers different views of a radiology scan. Based on this intuition, we further propose a co-training approach, where two machine learning models are built upon the \textit{Findings} and \textit{Impression} sections, respectively, and use each other's information to boost performance with massive unlabeled data in a semi-supervised manner. We conducted experiments in a public health surveillance study, and results show that our co-training approach is able to improve performance using the dual views and surpass competing supervised and semi-supervised methods. 
 \\ \newline 
 }

\begin{document}

\maketitleabstract

\renewcommand{\headrulewidth}{0pt}
\cfoot{\begin{myfont}In \textit{LREC-COLING 2024},~pages 477-483\end{myfont}}
\thispagestyle{fancy}
\fancyhead{}

\section{Introduction}
Radiology report analysis plays an important role in patient diagnosis and monitoring~\cite{diagnosis1,monitoring1,monitoring2}. For example, brain radiology reports---such as those derived from magnetic resonance imaging (MRI) and computed tomography (CT)---are typically in the form of free text, and can be used to determine the presence of brain tumors and track their progression over time. This helps collect surveillance data for public health initiatives~\cite{brain-surveillance}. 

Machine learning methods have been widely applied to the radiology domain, as the ever-growing volume of radiology reports makes it difficult for humans to label every single one. In early work, researchers perform manual feature engineering to construct classifiers such as decision trees~\cite{decisiontree} and support vector machines~\cite{svm}. More recently, deep learning has been a prevailing approach to radiology report analysis, leading to great advancements in the field. \citet{ALARM} finetune the BioBERT model~\citep{bioBERT} for MRI scan classification. \citet{CheXbert} use the labels produced by a traditional rule-based X-ray classifier~\citep{Chexpert} to train a BERT model~\citep{BERT}, which outperforms the rule-based classifier. 

However, we observe that existing methods do not make full use of the internal structures of a radiology report, which typically contains a \textit{Findings} section and an \textit{Impression} section. The former details factual observations made by a radiologist, whereas the latter synthesizes their findings into a summary~\citep{radling}. Our intuition is that such structural information can provide different views of a radiology report and improve the performance of machine learning systems.

In this paper, we propose a co-training approach to radiology report analysis, framing the  \textit{Findings} and \textit{Impression} sections as two different views. Specifically, we train two classifiers for \textit{Findings} and \textit{Impression}, respectively. Then, we use one classifier's predicted labels to train the other in a co-training fashion~\cite{cotrain}, which makes use of a large unlabeled dataset. These co-trained classifiers can be combined as an ensemble~\cite{ensemble} to make final predictions. In this way, the model trained on one section is able to glean information from the other in a semi-supervised manner. This allows us to make use of the structure of a typical radiology report as well as unlabeled data to improve overall performance.

We conducted experiments for a brain tumor surveillance project in collaboration with Alberta Health Services (AHS), a Canadian provincial health agency, where we are provided with de-identified historical radiology reports of real patients. The results show that co-training improves each individual model in a semi-supervised manner, and that their ensemble is able to further boost the performance. Our entire approach outperforms both supervised learning based on the small labeled data and self-train, a competing semi-supervised method.\footnote{Code available at:~\url{https://github.com/MANGA-UOFA/Radiology-Cotrain}}

% cite two papers for "gaining traction"
% cite another papers for "sup info"
\section{Related Work} 
\textbf{Semi-supervised learning} assumes only a small labeled dataset exists, and takes advantage of massive, readily available unlabeled data to improve model performance~\citep{ssl-survey}. Two popular frameworks are self-training and co-training. In self-training, a model generates pseudo-labels for the unlabeled data and trains itself~\citep{self-train-og}, whereas in co-training, two models are built on two views (different input information about a data sample) and co-train each other~\citep{cotrain}. In fact, co-training has been previously used in various NLP applications. \citet{cotrain-parse} use inside-span and outside-span views to co-train an unsupervised constituency parser; \citet{cotrain-mutual} use a query view and a document view to co-train a selective search system; and \citet{cotrain-prompt} use two different language models' representations to co-train and improve the performance of a prompting system. 

\textbf{Radiology report analysis} has gained traction in recent years~\citep{rad-classification,rad-extraction}, as the textual reports provide rich supplementary information to images~\cite{ALARM,multimodal-generation}. While an entire report can be the input to a machine learning system~\cite{Drozdov2020,full_report1}, researchers have realized the value of using the section structure of radiology reports. \citet{lymph-findings} use the \textit{Findings} section to extract information about lymph nodes in abdominal MRI reports; \citet{Chexpert} use the \textit{Impression} section to create a rule-based pathology classifier for chest X-rays. \citet{rad-summarization} train a text generation system to automatically synthesize an \textit{Impression} section from the \textit{Findings} section. 

To the best of our knowledge, we are the first to propose a co-training method based on \textit{Findings} and \textit{Impression}, as well as to build model ensembles of the two sections.

% For example, sombody does something. Somebody else does something else. % ideally,

%Specifically, NLP systems have been developed to create structural labels for radiology reports. 

% Previous work in this area have approached the structure of radiology reports in two main ways. One is to apply predictions on the entire radiology report~\citep{ALARM,full_report1,full_report2}. 
% The other is to choose a specific section of the radiology report to focus on~\citep{Chexpert,CheXbert,lymph-findings}. 

% In our work, we choose to use a co-training approach to perform radiology report analysis. Doing so allows us to harness the information in both the \textit{Findings} and \textit{Impression} sections.  

% \begin{figure*}[!t] 
%   \centering
%     \includegraphics[scale=0.17]{radiology_report.jpg}
%   \caption{The structure of a typical radiology report}
%   \label{fig:report}
% \end{figure*}

\section{Approach}
%In this section, we first give the problem formulation (Subsection \ref{subsect:problem-form}). Then, we sequentially introduce our approach to addressing this problem by first describing how we initialized the classifiers for the two views (Subsection \ref{subsect:initialization}), then we explain our co-train method (Subsection \ref{subsect:co-train}), and finally the inference strategy (Subsection \ref{subsect:ensemble}).  

\textbf{Formulation.}
Given a radiology report $\mathbf{x}$, our goal is to predict a label ${y} \in \{0, \cdots, K-1 \}$ with $K$ pre-determined categories. For example, an important label for brain radiology reports is  ${y} \in \{0, 1 \}$, indicating the absence or presence of a brain tumor.

In this work, we need to tackle a common and realistic setting for radiology report analysis: we only have a small set of labeled reports $\mathcal{D}_l = \{(\mathbf x^{(i)}, y^{(i)})\}^L_{i=1}$, but there exists a large unlabeled dataset $\mathcal{D}_u = \{\mathbf x^{(j)}\}^U_{j=1}$. 

\begin{figure*}[!ht] 
\begin{center} %\vspace{-0.2cm}
\includegraphics[width=\linewidth,scale=1]{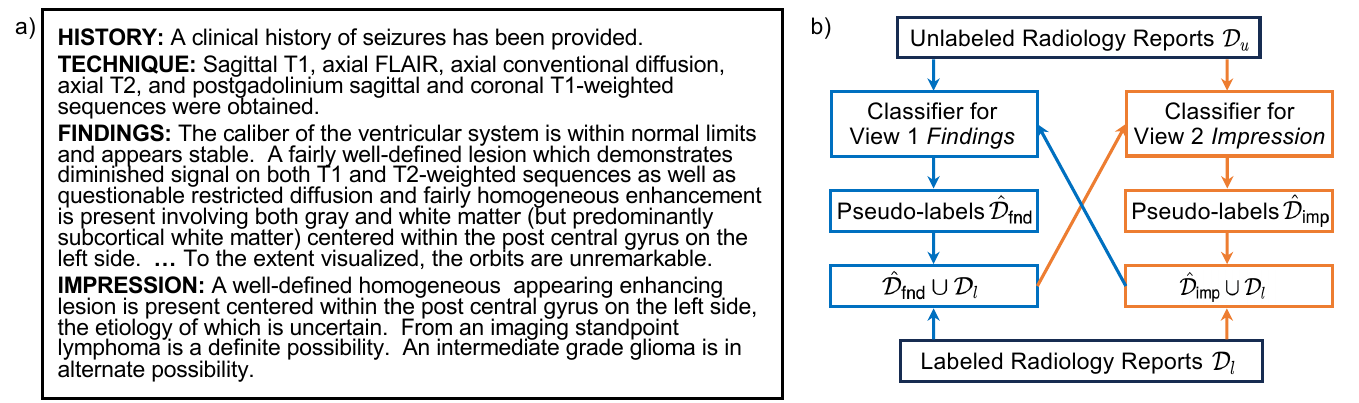} %\vspace{-0.7cm}
  \caption{a) A typical radiology report. b) An overview of our co-training approach.} %\vspace{-0.7cm}
  \label{fig:report_diagram}
\end{center}
\end{figure*}

Our intuition is that a typical radiology report has internal structures. In Figure~\ref{fig:report_diagram}a, for example, the report has several sections, namely, \textit{History}, \textit{Technique}, \textit{Findings}, and \textit{Impression}. In particular, we observe that \textit{History} does not provide the information of the current report and that \textit{Technique} explains the operational procedure; they are therefore not helpful for our task. On the other hand, the \textit{Findings} section describes all observations made by a radiologist, and the \textit{Impression} section summarizes and interprets the key findings. 
Thus, we discard \textit{History} and \textit{Technique} in our approach, but make use of \textit{Findings} (denoted by $\mathbf{x}_{\text{fnd}}$) and \textit{Impression} (denoted by $\mathbf{x}_{\text{imp}}$) as the two views for co-training. In other words, the input of a sample can be represented by $\mathbf x=(\mathbf{x}_{\text{fnd}}, \mathbf{x}_{\text{imp}})$.

\textbf{Supervised Initialization.}
Before co-training, we first use the small labeled data $\mathcal D_l$ to initialize the \textit{Findings} and \textit{Impression} classifiers, $P_\text{fnd}(y|\mathbf x_\text{fnd};\bm\theta_\text{fnd})$ and $P_\text{imp}(y|\mathbf x_\text{imp};\bm\theta_\text{imp})$.  Specifically, we first train them by finetuning  DistilBERT~\cite{distilBERT}, a small distilled version of the pretrained language model BERT~\cite{BERT}.\footnote{We chose the DistilBERT model because regulations require that the research has to be conducted within the server provided by AHS, which has limited memory.} Take the \textit{Findings} view as an example: we apply linear transformation to the final layer's hidden state associated with [CLS], a token prepended to a sequence for classification. Then, a softmax function predicts a probability distribution\footnote{We slightly misuse the notation that $P(y|\cdot)$ refers to a probability distribution, where $y$ is a random variable taking values in $\{0,\cdots, K-1\}$ for $K$-way classification.} by
$P(y|\mathbf x_\text{fnd};\bm\theta_\text{fnd})=\operatorname{softmax}(\bm W_\text{fnd}\bm h_\text{[CLS]}+\bm b_\text{fnd})
$, where $\bm h_\text{[CLS]}$ is the [CLS] token's representation at the last hidden layer. $\bm \theta_\text{fnd}$ is the entire parameter set, including softmax-layer parameters ($\bm W_\text{fnd}$ and $\bm b_\text{fnd}$), as well as the parameters of the \textit{Findings}-view model.
The training is accomplished by maximum likelihood estimation with labeled data $\mathcal D_l$:
\begin{align}
\bm \theta_{\text{fnd}} = \operatorname*{argmax}\nolimits_{\bm \theta_\text{fnd}} \sum\nolimits_{i=1}^L \operatorname{log}P(y^{(i)}|\mathbf  x^{(i)}_{\text{fnd}}; \bm \theta_{\text{fnd}})
\end{align}

Likewise, we train a classifier $P_\text{imp}(y|\mathbf x_\text{imp};\bm\theta_\text{imp})$ for the \textit{Impression} view. These supervised classifiers serve as a good starting point for our co-training procedure. 

\textbf{Co-Training.}
The overview of our approach is presented in Figure~\ref{fig:report_diagram}b. We maintain two classifiers for \textit{Findings} and \textit{Impression}, respectively. 
Our co-training approach alternately applies each classifier to the unlabeled data $\mathcal D_u$, which produces pseudo-labels to co-train the other classifier. This process is repeated for performance improvement.

Consider using the classifier for \textit{Findings} to co-train that for \textit{Impression}.  For every unlabeled sample $\mathbf x^{(j)}$ in $\mathcal D_u$, we apply the \textit{Findings} classifier and obtain its prediction
\begin{align}
\hat{y}^{(j)}_{\text{fnd}}=\operatorname*{argmax}\nolimits_y P(y|\mathbf x_\text{fnd}^{(j)};\bm\theta_\text{fnd})
\end{align}
with its predicted probability $P(\hat{y}^{(j)}_{\text{fnd}}|\mathbf x_\text{fnd}^{(j)};\bm\theta_\text{fnd})$. 

Then, we add high-quality labels to the training set based on two criteria. First, we select samples for which the two classifiers agree, in order to avoid confusion during co-training. That is, we have $\hat{y}_{\text{fnd}}^{(j)} = \hat{y}_{\text{imp}}^{(j)}$, where $\hat{y}^{(j)}_{\text{imp}}=\operatorname*{argmax}_y P(y|\mathbf x_\text{imp}^{(j)};\bm\theta_\text{imp})$. Second, we choose the samples with top-$k\%$ of the  \textit{Findings}-predicted probabilities among the agreed labels. This is based on the intuition that labels with higher probabilities are more likely to be correct~\cite{cotrain,self-train-og}, which further ensures the quality of our pseudo-labels.  Overall, our pseudo-labeled dataset has the form of 
$\hat{\mathcal D}_{\text{fnd}}= \operatorname{top-\!}k\%\{(\mathbf x^{(j)}, \hat{y}_\text{fnd}^{(j)}): \hat{y}_{\text{fnd}}^{(j)} = \hat{y}_{\text{imp}}^{(j)}, \mathbf x^{(j)}\in\mathcal D_u\}$, which is merged into the labeled one as $\hat{\mathcal D}_{\text{fnd}}\cup \mathcal D_l$ to train the \textit{Impression} classifier. 

The roles then reverse to re-train the \textit{Findings} classifier $P_\text{fnd}(y|\mathbf x_\text{fnd};\bm\theta_\text{fnd})$ 
using the \textit{Impression}-predicted pseudo-labels along with the original small labeled data, given by $\hat{\mathcal D}_{\text{imp}}\cup \mathcal D_l$ .  Co-training continues in a such a way until validation performance peaks. 

This framework allows for two views of a radiology report: the detailed factual observations in the \textit{Findings} section and the concise synthesized information in the \textit{Impression} section. Together, they can help each other during the co-training process and improve the classification performance. 

\textbf{Ensemble for Inference.}
To perform inference, we combine the co-trained classifiers by an average ensemble~\citep{ensemble}. Given an unseen radiology report $\mathbf x^{*}=(\mathbf{x}^{*}_{\text{fnd}}, \mathbf{x}^{*}_{\text{imp}})$, we apply both the
\textit{Findings} and \textit{Impression} classifiers to the respective sections and choose the most likely category based on averaged predicted probabilities:
\begin{align}\nonumber
\hat{y}^{*}=\operatorname*{argmax}\nolimits_y \tfrac{1}{2}\big [P(y|\mathbf x_\text{fnd}^{*};\bm\theta_\text{fnd}) + P(y|\mathbf x_\text{imp}^{*};\bm\theta_\text{imp})\big]
\end{align}

\begin{table}[!t] 
	\centering
     %\vspace{0cm}
	\resizebox{0.68\linewidth}{!}{
	\begin{tabular}{lccc}
	\toprule
		Task $\backslash$ Label&0& 1& 2\\
	\midrule
        BT& 331 & 537 & -- \\
        \midrule
        Aggressiveness & 331 & 344 & 193\\
        \bottomrule
        \end{tabular}}
    %\vspace{-.2cm}
	\caption{Label distribution in each task. BT and Aggressiveness labels do not necessarily agree, and the number of 331 is a coincidence.} \label{tab:dataset}
    %\vspace{-.5cm}
    \end{table}

The ensemble approach makes use of the two views (\textit{Findings} and \textit{Impression}) by smoothing out the noise of the individual classifiers.

\begin{table*}[!t]
\centering
\resizebox{0.8\linewidth}{!}{
\begin{tabular}{>{\centering\arraybackslash}p{0.2\textwidth}|>{\centering\arraybackslash}p{0.05\textwidth}|>{\centering\arraybackslash}lll}
\hline
Setting & Row & Model & BT & Aggressiveness \\ 
\hline
 & 1 & Concat & 0.8837 $_{(\text{0.8832}\pm{\text{0.009}})}$ & 0.8330 $_{(\text{0.8270}\pm{\text{0.019}})}$  \\
 & 2 & Findings & 0.8825  $_{(\text{0.8833}\pm{\text{0.012}})}$  & 0.7546 $_{(\text{0.7768}\pm{\text{0.018}})}$ \\
 & 3 & Impression & 0.9102 $_{(\text{0.8915}\pm{\text{0.013}})}$ & 0.8445 $_{(\text{0.8549}\pm{\text{0.007}})}$ \\ 
\multirow{-4}{*}{\begin{tabular}[c]{@{}@{}@{}c@{}@{}@{}}Supervised\\ (520 labeled\\samples)\end{tabular}} & 4 & Ensemble (2 + 3) & 0.9148 $_{(\text{0.9122}\pm{\text{0.007}})}$ & 0.8676 $_{(\text{0.8646}\pm{\text{0.012}})}$ \\
\hline
 & 5 & Concat (self-train) & 0.8952 & 0.8563 \\
 & 6 & Findings  (self-train) & 0.8906 & 0.7845 \\
 & 7 & Impression  (self-train)  & 0.8860 & 0.8560 \\
 & 8 & Ensemble (6 + 7) & 0.9160 & 0.8802 \\
\cline{2-5}
 & 9 & Findings  (co-train) & 0.8906 & 0.8399 \\
 & 10 & Impression  (co-train) & 0.9044 & 0.8621 \\
\multirow{-7}{*}{\begin{tabular}[c]{@{}@{}@{}c@{}@{}@{}}Semi-Supervised\\ (+10K unlabeled \\samples)\end{tabular}} & 11 & Ensemble (9 + 10) & \textbf{0.9286} & \textbf{0.8848} \\  
\hline
\end{tabular}}    %\vspace{-.2cm}
\caption{Accuracy on the Brain Tumor (BT) and Aggressiveness classification tasks. ``Concat'' refers to training a single model with concatenated \textit{Findings} and \textit{Impression} sections (without the section titles) as the input. 
For the supervised setting, we trained the model with 5 different seeds, and report the median and the (mean$\pm$standard deviation). Due to limited computing resources, self-training and co-training settings were run once initialized from the median run of the supervised setting.}
\label{tab:main_results}    %\vspace{-.4cm}
\end{table*}

\section{Experiments}
\subsection{Setup}
We evaluated our approach for a project collaborated with Alberta Health Services, where the goal is to improve surveillance for brain tumors with historical textual radiology reports, including both CT and MRI scans. To reach this goal, the project focuses on two important labels:

$\bullet$ \textbf{Brain Tumor (BT):} The classification goal is ${y} \in \{0, 1 \}$ indicating whether the radiology report suggests there is one or more brain tumors observed in the scan. %any standard survey/review of PBT

$\bullet$ \textbf{Aggressiveness:} Here, the classification goal is ${y} \in \{0, 1, 2 \}$ referring to non-aggressive, aggressive, or possibly aggressive, respectively. Notice that the aggressiveness label provides different information from BT, which can be either aggressive or non-aggressive; on the other hand, an aggressive label can also be a cancer metastasis (cancer spread) that has no tumor, e.g., leukemia spread into the brain~\cite{leptomeningeal}. 

The dataset contains 868 radiology reports, manually annotated with the above two labels of interest, as well as 10K unlabeled radiology reports. Each report has a \textit{Findings} section and \textit{Impression} section, containing on average 219 and 55 tokens, respectively. The label distribution of each task is shown in Table~\ref{tab:dataset}. In this paper, we treat the two labels as independent tasks for the evaluation of our approach. 

\textbf{Implementation Details.}
Due to the small dataset, we performed 5-fold cross-validation for robust evaluation. We split the entire dataset into five folds, and for the test of each fold (174 samples), we used three folds for training (520 samples) and the other fold for validation (174 samples).

We used the AdamW optimizer~\citep{AdamW} with a batch size of 16 and a standard learning rate of 5e-5~\citep{BERT}. Early stopping was implemented based on validation performance in each co-training round; we set the number of maximum co-training rounds to be 5, also early stopped by validation. For the BT task, the top-50\% pseudo-labels were added at each co-training step, whereas for the Aggressiveness task, the top-25\% were added. We will analyze the choice of top-$k$\% in \S\ref{subsect:results}.
 
\subsection{Results}
Table~\ref{tab:main_results} shows the main results on the two tasks, BT and Aggressiveness. Here, the performance is measured by accuracy, since our dataset is relatively balanced as shown in Table~\ref{tab:dataset}.

We first analyze the use of \textit{Findings} and \textit{Impression} sections in the supervised setting only (\mbox{Rows~1--4}). As seen, \textit{Impression} (Row~3) yields higher performance than \textit{Findings} (Row~2), especially for the Aggressiveness task. This is understandable because \textit{Impression} is a synthesized summary and better aligns with the actual label. Concatenation of the two sections, without section titles (Row~1) does not outperform \textit{Impression} only  (Row~3), as the \textit{Findings} section may be of considerable length and confuse the model. Our ensemble approach, even without co-training (Row~4), achieves consistent improvement on both tasks, justifying our dual views that make use of the internal structure of a radiology report. 

Then, we performed semi-supervised learning using 10K unlabeled samples. In addition to our co-training method, we experimented with a self-training baseline, where a model uses its own predictions to boost its performance. In general, semi-supervised learning (Rows~5--11)  outperforms small-scale supervised learning (Rows~1--4), except for a minor inconsistency of \textit{Impression} on the BT task. The results verify that massive unlabeled data can alleviate the data sparsity problem in the medical domain.

\begin{figure}[!t] 
\begin{center}    \vspace{-.2cm}
\includegraphics[scale=0.1]{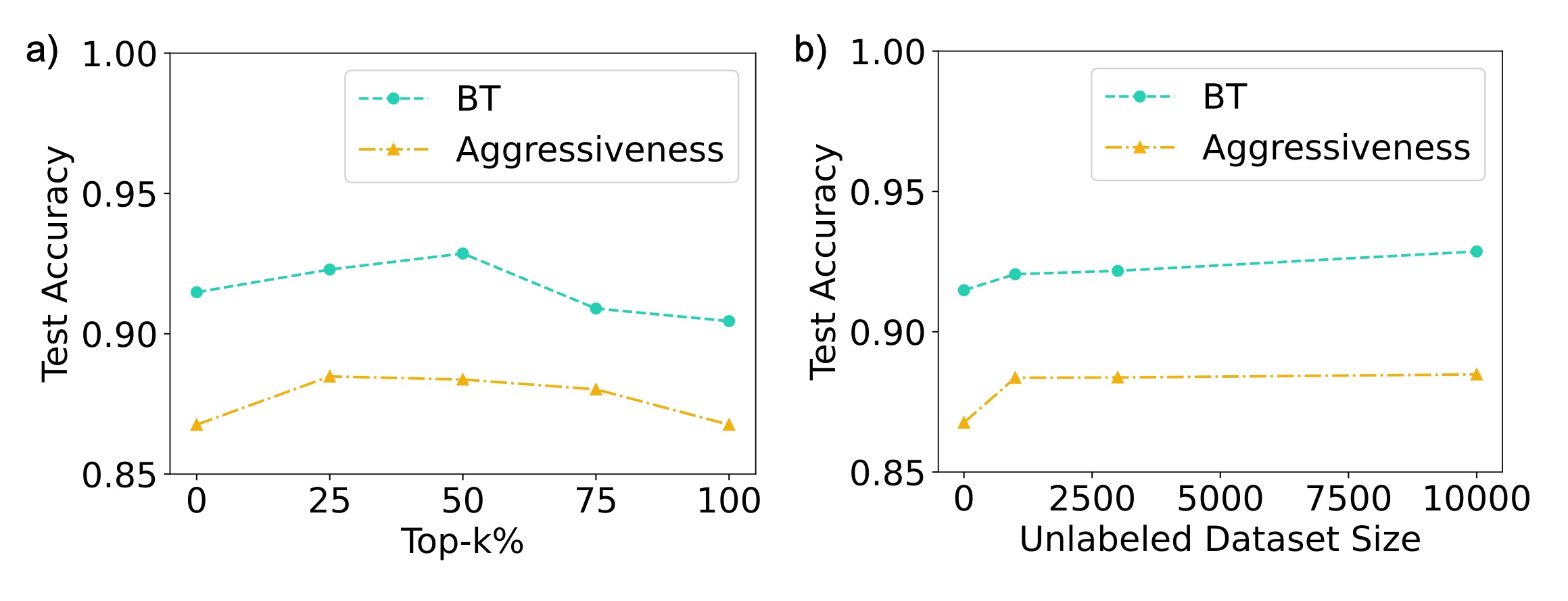} \vspace{-0.5cm}
  \caption{a) Effect of top-k$\%$ on co-training performance, when fixing the unlabeled dataset size at 10,000. b) Effect of unlabeled dataset size, using $k=50$ for BT and $k=25$ for Aggressiveness.}
  \label{fig:topk} \vspace{-.4cm}
\end{center}
\end{figure}
Among semi-supervised learning methods, we observe that co-training always excels when compared with self-training. For example, our co-trained \textit{Findings} classifier (Row~9) improves the accuracy by 8 points compared with supervised learning (Row~2) on Aggressiveness, whereas the self-trained \textit{Findings} classifier (Row~6) is only improved by 3 points. Our ensemble (Row~11) is able to further boost the performance and surpasses the ensemble of supervised counterparts (Row~4).  This indicates that exchanging dual-view information by co-training is more powerful than a \textit{post hoc} ensemble.

Moreover, our full method (co-training and ensemble, Row~11) outperforms a na\"ive application of DistilBERT to radiology reports (Row~1) by 4.49 and 5.18 percentage points for BT and Aggressiveness, respectively. This significant jump in performance further suggests that our approach is effective. 

Overall, the experiments convincingly show that making use of the \textit{Findings} and \textit{Impression} sections benefit radiology report analysis, and that our co-training approach is able to strategically exploit the dual views of the report to gain additional benefits.
~\label{subsect:results}

\textbf{Detailed Analyses.} In our approach, we choose top-$k\%$ confident samples for co-training, and we analyzed the effect of the hyperparameter $k$ in Figure~\ref{fig:topk}a. As seen, a modest $k$ yields highest performance, which is reasonable because a smaller $k$ results in fewer pseudo-labels, whereas a larger $k$ brings in more noise. Based on the analysis, we chose $k=50$ for BT and $k=25$ for Aggressiveness. 

We also analyzed the role of the unlabeled dataset size in the co-training process. Figure~\ref{fig:topk}b shows that the performance is generally improved with more data, but is not sensitive to the exact number of samples. We chose 10K unlabeled samples for co-training in our experiments.

\section{Conclusion}
In this paper, we propose a co-training approach to radiology report analysis, where we regard the \textit{Findings} and \textit{Impression} sections as dual views of a radiology report. We conducted experiments on two tasks: Brain Tumor (BT) classification and Aggressiveness classification. The experimental results demonstrate that our co-training method is able to make use of the dual views with unlabeled data in a semi-supervised manner, and outperforms different competing methods. We further provide detailed analyses of our proposed co-training method.

\section{Ethics Statements}
Our study involves de-identified patient data and human annotations. We have obtained ethics approvals from our research institute, as well as the partner government agency of public health. This project will provide timely and accurate data that can be used to inform health system resource allocation and improve understanding of the cancer recurrence/progression at the population level. 

\section{Acknowledgments}
The research is supported in part by the Canadian Cancer Society, the Natural Sciences and Engineering Research Council of Canada (NSERC), the Amii Fellow Program, the Canada CIFAR AI Chair Program, an Alberta Innovates Project, a UAHJIC Project, a donation from DeepMind, and the Digital Research Alliance of Canada (alliancecan.ca).

\newpage
% \vspace{.1cm}
\nocite{*}
\section{Bibliographical References}\label{sec:reference}

\vspace{-.7cm}
\bibliographystyle{lrec-coling2024-natbib}
\bibliography{lrec-coling2024-example}

\end{document}